# Multi-Scenario Reasoning: Unlocking Cognitive Autonomy in Humanoid Robots for Multimodal Understanding

Libo Wang

*Abstract*— **To improve the cognitive autonomy of humanoid robots, this research proposes a multi-scenario reasoning architecture to solve the technical shortcomings of multi-modal understanding in this field. It draws on simulation based experimental design that adopts multi-modal synthesis (visual, auditory, tactile) and builds a simulator "Mahā" to perform the experiment. The findings demonstrate the feasibility of this architecture in multimodal data.**

## I.INTRODUCTION

In tandem with the support of cutting-edge GPUs, artificial intelligence technologies such as model predictive control, multi-sensor fusion, visual SLAM, and sim-to-real continue to stimulate the development potential of humanoid robots (Makoviychuk et al. , 2021; Chen et al., 2024; Dao et al., 2024). Given the previous literature on the development of cognitive architecture as a solution to human-level autonomy, cognitive autonomy is still a major problem in the thinking, planning and action selection modules of humanoid robots (Ogunsina et al. al., 2024). For example, Burghart et al (2005) have demonstrated a three-level cognitive architecture that from environmental input to task execution (Fig. 1).

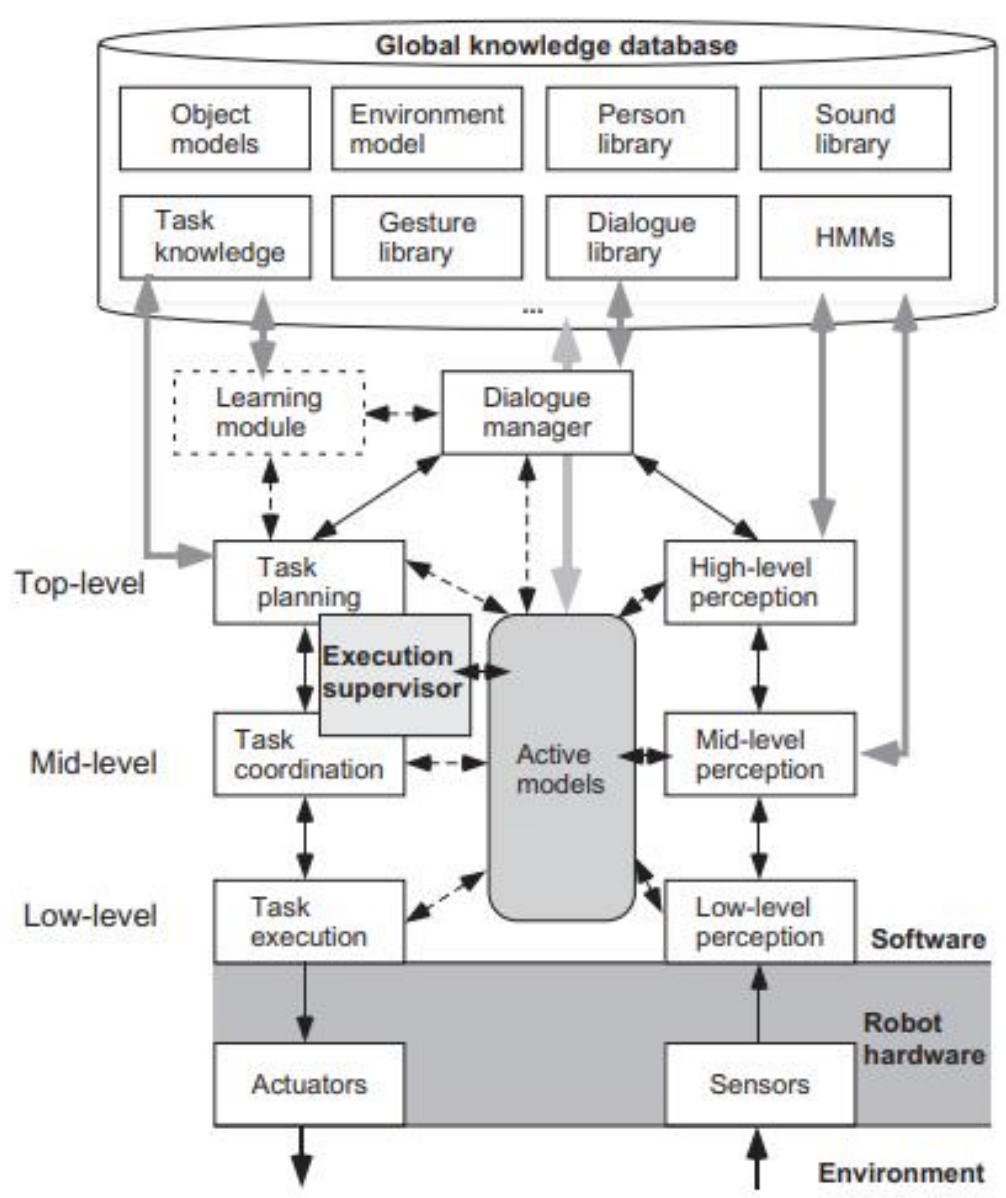

Fig. 1 Cognitive architecture of the Karlsruhe Humanoid Robot (Adapted from Burghart et al., 2005)

However, current technology supports humanoid robots to imitate humans to perform repetitive operations, but it does not mean that the simulation reaches human-level cognitive autonomy (Katiyar & Katiyar, 2021; Ogunsina et al., 2024). The perception and execution capabilities cannot make up for the significant challenges in human-level autonomy that are reflected in cognitive abilities such as thinking, planning, and decision-making (Ogunsina et al., 2024). Target perception, tracking, and repetitive movement planning still rely on visual processing (Guo et al., 2020). Humanoid robots are not only based on pre-trained architectural models of static data, but also need to process multi-modal data (visual, auditory, tactile) and flexible and coherent reasoning (Fisher et al., 2021; Xiao et al., 2023).

The technical shortcomings of multimodal understanding are the key gaps that make it difficult for humanoid robots to achieve cognitive autonomy. Multimodal understanding requires extracting relevant features from different sensory inputs and integrating them into a consistent semantic representation through cross-modal alignment techniques (Duan et al., 2022; Navarro-Guerrero et al., 2023; Tong et al., 2024). However, in practice, it is difficult to efficiently integrate and process multi-modal data such as visual, auditory and tactile data, resulting in ambiguous semantics and incoherent responses.

As mentioned before, architectures that rely on static pre-training data to complete tasks lack the ability to integrate cross-modal data (Ye et al., 2023). This directly causes semantic ambiguity when humanoid robots process feedback with hearing or touch or response incoherence (Pramanick & Rossi, 2024). Although some research attempts multi-modal fusion technology, progress is still limited and is not enough to provide humanoid robots with the same adaptive capabilities as humans (Yuan et al., 2024). To address this gap, this research proposes a multi-scenario reasoning architecture as an innovative solution. It aims to leverage multi-scenario reasoning to optimize the key challenges of cognitive autonomy in humanoid robots' cross-modal understanding of visual, auditory and tactile data based on current technology shortcomings.

## II.MULTI-SCENARIO REASONING

### A. Theoretical Foundation

The principle of multi-scenario reasoning proposed in this research and applied to humanoid robots is inspired by situated cognition theory. This theory emphasizes that the environment is inextricably linked to the behavior of knowledge generation and application, and that it embodies meaning through individual interactions in real situations (Jenlink & Austin, 2013).

Specifically, reasoning ability is affected by the real-time integration of multi-modal sensory information from the five senses of vision, hearing, smell, taste, and touch and the behavior of combining situation selection (Wang et al., 2003;

Thagard, 2010). The human brain's response to situations The construction and integration process of semantics is carried out in the prefrontal cortex, temporal lobe, and parietal lobe (Jouen et al., 2015). The memory retrieval and contextual association of the hippocampus play an important role in processing environmental information for cross-modal reasoning in multiple scenes (Morici et al., 2022). This process involves the dynamic allocation of attention resources to ensure the human brain's immediate understanding of complex situations and selected optimal actions, which profoundly affects the operation process of the multi-scenario reasoning architecture designed in this research (Nicolini et al., 2024).

As mentioned before, the principle of multi-scenario reasoning is inspired by situated cognition theory, thereby extending its application to the cognitive autonomy design of humanoid robots. Different from traditional technologies that only focus on static scenes or single modal data, it imitates the principle of the human brain to instantly integrate multi-modal data in complex situations to achieve dynamic processing of visual, auditory and tactile information of humanoid robots across scenarios.

### B. *Conceputal Principles*

As a concept proposed by this research to improve the cognitive autonomy of robots, multi-scenario reasoning has been applied to the thinking, planning and decision-making fields of humanoid robots for the first time. It focuses on semantic integration and synchronization of visual, auditory and tactile data from external sensors to make optimal action choices in different dynamic scenarios. It builds a multi-scenario reasoning architecture to simulate real-time processing of multi-modal data and improves the dynamic adaptability of adaptive systems in cross-scenario learning.

Based on the principle of situated cognition theory, multi-scenario reasoning simulates the semantic integration and decision-making capabilities of the human brain's reasoning process in an uncertain environment. In the field of humanoid robots, the application of multi-scenario reasoning focuses on global situation modeling based on multi-modal data, and uses this to perform continuous reasoning and dynamic adjustment. Multi-scenario reasoning promotes cognitive autonomy in complex environments through semantic alignment, synchronized processing and scenarios of multi-modal data during thinking, planning and decision-making.

The core principle of multi-scene reasoning lies in dynamic scene modeling and semantic integration of multi-modal data, thereby solving the key shortcomings of existing multi-modal understanding technology. Based on the principle of situated cognition theory, this concept simulates the semantic integration and decision-making of the human brain in an uncertain environment. From a technical practice perspective, it is based on semantic alignment and uses situational analysis to uniformly represent visual, auditory and tactile data, which optimizes the robot's cross-modal understanding capabilities.Sparse attention effectively optimizes and dynamically adjusts the weight of each modality to highlight key information, thereby improving the accuracy and efficiency of multi-modal data processing (Song et al., 2024). The memory-augmented module provides the ability to trace back data in long-term and dynamic scenarios, ensuring that robots can perform efficient reasoning based on historical context in real-time scenarios (Muthirayan et al., 2020). The scene reasoning results are entity mapped through the sim2real module to promote the seamless connection of the actual operation of the robot (Zhao et al., 2023).

## III. Proposed Architecture & Algorithms

This section is closely connected with the previous theories and concepts to explain in detail the core modules and algorithm display of the multi-scenario reasoning architecture. Fig. 2 explains modules including data input, scenario processing, attention-based prioritization, memory-augmented reasoning, action-decision modeling, sim2real and selected optimal action.

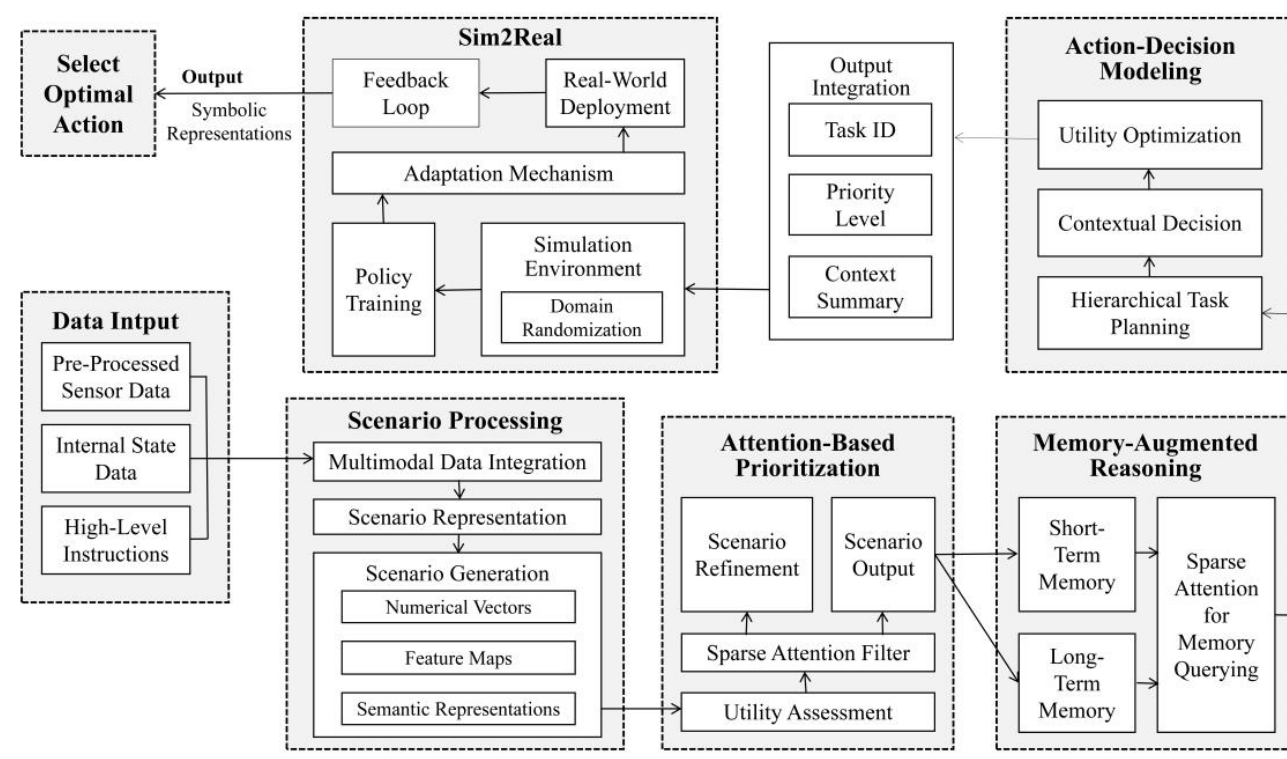

### A. *Abbreviations and Acronyms*

This module is responsible for preprocessing and structuring data from multi-modal perceptrons, and providing high-quality data sources to subsequent modules. The Data input contains three subcomponents. Among them, pre-processed sensor data is used to integrate visual, auditory and tactile data; pnternal state data focuses on processing the internal status information of humanoid robots; high-level instructions are used to parse high-level instructions and generate preliminary semantic data.

The following algorithm is divided into four parts: data reprocessing, multimodal normalization, feature extraction, integrated multimodal output. $D$ is the raw data set; $T(d)$ is a trustworthiness function; $\tau$ is the threshold. Normalizes data $D(m)$ for each modality using mean $\mu(m)$ and standard deviation $\sigma(m)$. $F_m$ is extracted features. $m \in$ {visual, auditory, tactile}. Combines extracted features from all modalities into a unified output $O$.

$$D_{filtered} = \{d \in D \mid T(d) > \tau\}$$
$$D_{norm}(m) = [D(m) - \mu(m)] / \sigma(m),$$
$$F_m = \text{Extract}(D_{norm}(m))$$
$$O = \bigcup_m F_m$$

### B. *Scenario Processing*

This module performs semantic modeling and scene representation for multi-modal data to ensure data consistency and adaptability in dynamic environments. In components, multimodal data integration for semantic integration of cross-modal data; scenario representation converts data into structured representations for robot processing; scenario generation builds scenarios based on feature maps and semantic labels, and generates semantic consistency through semantic representations data output.

The algorithm of this module supports functions such as multimodal data integration, feature mapping, scenario generation and evaluation, and sparse attention-based scenario selection. Notably, $R_k$ is a semantic feature, $U_k$ is data point.

- Input:

$S$, $I$, $H$, weights $\alpha_s$, $\alpha_i$, $\alpha_h$.
Unified data vector $U = [U_1, U_2, ..., U_n]$
Feature map $M$
Scenarios $S = \{S_1, S_2,...,S_m\}$, utilities $U(S_j)$, selection size $k$.

- Multimodal Data Integration

$$U_k = \alpha_s s_k + \alpha_i i_k + \alpha_h h_k, \quad \forall k \in \{1, 2, ..., n\}$$

- Feature Mapping

$$R_k = f(U_k),\ f(U_k) = \text{round}\ (U_k, 2)$$
$$M_k = g(U_k),\ g(U_k) = e^{-U_k}$$

- Scenario Generation

$$S_{j,k} = M_k + \Delta(M_k),\ \Delta(M_k) \sim U(-0.1, 0.1)$$

Compute scenario utility:

$$U(S_j) = \Sigma_{k=1}^{n} S_{j,k}$$

- Sparse Attention-Based Scenario Selection

Select top-k scenarios:

$$S_k = \text{Top - k}(U(S_i)),\ \ \forall_j \in \{1, 2, ..., m\}$$

- Output: Selected scenarios $S_k$.

*C. Scenario Processing*

The attention-based prioritization module is designed to dynamically filter and weight multi-modal data to optimize inference accuracy and efficiency. Sparse attention filter technology is used to filter out key data; Scenario refinement ensures semantic clarity during the optimization of scene output; Utility assessment evaluates data priority based on scene semantics; output scenario is responsible for generating these outputs.

In algorithms, $r(S_j)$ is the relevance score for $S_j$. Top-k($\cdot$) identifies the k scenarios with the highest scores; $S'_i$ is the refined scenario. $f(\cdot)$ is a function incorporating memory-based adjustments. $a'_j$ is adjusted attribute of the scenario; $g(\cdot)$: Adjustment function based on memory $m_j$. $\mu(m)$ represents mean, $\sigma(m)$ represents standard deviation.

- Input

Raw data $D$, consisting of multimodal input $\{d_1, d_2,..., d_n\}$.
Trustworthiness function $T(d)$ for evaluating data quality.
Threshold $\tau$ for data filtering.
Modalities $m \in$ {visual, auditory, tactile}.

- Sparse Attention Filtering

$$r(S_i) = \exp\ (U(S_i)) / \Sigma^{m}_{k=1} \exp(U(S_k))$$
$$S_k = \text{Top - k}(r(S_j))$$

- Scenario Refinement

$$a'_j = g\ (a_j, m_j)$$

- Output

Refined top $k$ scenarios $S'_k = \{S'_1, S'_2, ..., S'_k\}$

*D. Memory-Augmented Reasoning*

This module introduces memory-augmented network to improve the accuracy and coherence of reasoning. Drawing on short-term memory to record the data of the current scene to support instant reasoning, long-term memory saves historical data for cross-scenario correlation analysis. Sparse attention for memory querying retrieves key data through attention screening technology. It simulates dynamic interaction with attention-based prioritization and outputs the reasoning results to the action-decision modeling module for strategy generation (Lou et al., 2024).

These algorithm explains that STM($t$) is spdated short-term memory at time t; ΔScenario is new contextual information from current input; $LTM_{relevant}$ is retrieved memory entry most similar to the query; Sim($\cdot$) is similarity function; $LTM_j$ is historical memory entries. $A_i$ is attention weight for memory entry $i$; Query is input query vector; $Memory_i$ includes STM or LTM; Output is weighted sum of relevant memory entries.

- Input

Query vector: Query.
Memory entries: Memory = {STM, LTM}.
Historical memory entries $LTM_i$.

- Operations

$$STM(t) = STM\ (t - 1) + \Delta Scenario$$
$$LTM_{relevant} = \arg\max\ (Sim(Query, LTM_i))$$
$$\alpha_i = \exp\ (Score(Query, Memory_i))\ /\ \Sigma_j \exp\ (Score(Query, Memory_j))$$
$$Score\ (Query, Memory_i) = (Query \cdot Memory_i)\ /\ (\|Query\|\ \|Memory_i\|)$$

- Output = $\Sigma_i\ \alpha_i \cdot Memory_i$

*E. Action-Decision Modeling*

The action-decision modeling module develops optimal action strategies based on the results of memory reasoning. Among them, utility optimization is responsible for calculating action utility; Contextual decision adjusts dynamics based on contextual information; action strategy is decomposed into specific executable steps by hierarchical task planning, and output integration is performed through task ID, priority level, and context summary.

In the following algorithm, $h_i$ is a subtask derived from task $t$; U is utility based on constraints C; $w_i$ is weights for context factors $C_i$; λ adjusts the impact of previous outcomes. $D$ is decisions

- Input

High-level task $T$.
Environmental context $E$.
Context factors $C_i(D)$.
Historical feedback data.

- Operations

$$H(t) = \{h_1, h_2, ..., h_n\}$$
$$P(h_i) = U\ (h_i|C)$$
$$D_{opt} = \arg\max\ Utility(D|E)$$
$$Utility(D|E) = \Sigma^{n}_{i=1}\ w_i \cdot C_i(D)$$

$$U(D) = \text{Predicted Outcome}(D) + \lambda \cdot \text{Historical Feedback}(D)$$

*F. Sim2Real*

Sim2real is responsible for mapping the planning plans generated by action-decision modeling in the simulation environment to real scenarios to achieve from reasoning to selected optimal action. The simulation environment in this module is used to simulate dynamic scenarios; policy training aims to strengthen the adaptability of behavioral strategies; adaptation mechanism is dedicated to mapping simulated behaviors to real environments; domain randomization enhances generalization capabilities; Real-world deployment implements it in practice.

These components draw on algorithms by Clavera et al (2018), Luo et al. (2018), Pateria et al. (2021), Moerland et al. (2023) and Zhang et al (2024). Specifically, $\varepsilon$ is the randomized environment; $E$ is base environment; $\mu$ and $\sigma$ define distribution of variations, $P$ is the probability distribution governing randomization; $\pi$ is the policy, $\tau$ represents trajectories in the simulation; $R(s_t, a_t)$ is reward for state $s_t$ and action $a_t$, $\gamma$ is the discount factor; $\varphi$ is the feature encoder; $D$ is the discriminator distinguishing between simulation and real-world data; $L_{adv}$ is the adversarial loss, $L_{task}$ isthe task-specific loss; $R_{real}$ represents real-world; $R_{sim}$ represents simulated rewards; $\delta$ is the reward discrepancy; $\alpha$ is adjustment factor.

- Input

Base environment $E$, including environmental parameters.
Simulation trajectories $\tau$ with states $s_t$ and actions $a_t$.
Real-world feedback: $R_{real}(s, a)$.
Simulated rewards: $R_{sim}(s, a)$.
Hyperparameters: $\mu, \sigma, P, \gamma, \alpha, \lambda$.

- Operations

$$\varepsilon = \text{Randomize}(E; \mu, \sigma, P)$$
$$\pi^* = \arg\max E_{\tau\sim\pi}\left[\Sigma^{T}_{t=0}\, \gamma^t R(s_t, a_t)\right]$$
$$min_{\varphi}\, max_D\, L_{adv}(\varphi, D) + \lambda\, L_{task}(\varphi)$$
$$\delta = R_{real}(s, a) - R_{sim}(s, a)$$
$$\pi^+ = \arg\max_{\pi}\left[\Sigma^{T}_{t=0}\, \gamma^t (R_{real}(s_t, a_t) + \alpha\delta)\right]$$

*G. Selected Optimal Action*

Serving as the output module of this architecture, selected optimal action integrates the results generated by sim2real to execute optimal action instructions. Through dynamic analysis and decision-making optimization, this module ensures the accuracy of action selection performed by the humanoid robot and optimizes scene processing with feedback data.

## IV. Experimentation

This research draws on the principle of simulation based experimental design in the category of quantitative experimental research methods to test the effectiveness of the multi-scenario reasoning architecture in solving the shortcomings of multi-modal understanding technology (Ekren et al., 2010; Saglam & Papelis, 2024). Given that many development institutions such as Tesla, Boston Dynamics and NVIDIA currently do not open source technical details in the field of thinking, planning and decision-making of humanoid robots, the objectivity of real experiments is challenged. To this end, the researcher adopted a single-group design to focus on the detection and evaluation of architecture performance.

*A. Experimentation Setup*

The researcher used code (uploaded to Github) and prompt engineering to train an experimental tool called Mahā based on a custom GPTs model that is used to simulate humanoid robots to perform multi-scenario reasoning. Developed by Meta, Sapiens-2B is primarily intended for high-resolution tasks centered around human vision, reasoning that the applicability of synthetic data is not just visual. In contrast, Mahā is more suitable as a preferred research tool that serves multi-scenario reasoning architectures. Another advantage of Mahā based on LLMs is that there is no need to consider hardware configuration, and the data synthesis method is simple and practical.

*B. Dataset*

Based on the objectivity of simulation experiments, the design is selected to generate synthetic data to satisfy the sparse attention filter and memory-augmented reasoning based on scene decision-making and prioritization. Given the above agency restrictions on publicly available training data, employing synthetic data increases legal and ethical freedoms. In addtion, synthetic data is not unique to this research, it has been integrated into research in related fields such as training and testing robots (Kim et al., 2024).

Motivated by the need for more advanced multimodal LLMs to integrate text, images, and sensor outputs, the researcher used Gemini 2.0 Experimental Advanced to generate visual, auditory, and execution multimodal synthetic data through prompt engineering.

*C. Implementation*

To ensure that it can be understood and executed by Mahā, the multi-modal synthetic data created through the sample code generation loop are all based on precise computer language. The researcher executed the code generated by Gemini 2.0 Experimental Advanced in Python 3.13 IDLE. This code obtained the JSON document as the data set for this research after iterating 10,000 times. The researcher used consistent prompts in the design of the running process in each modality to facilitate objective recording. To ensure the objectivity of multi-scenario reasoning ability assessment, the same piece of synthetic data can only be executed once in Mahā.

Since the current OpenAI knowledge base is updated to October 2023, Mahā may still be based on GPT 4-Turbo and has limited computing power. The researcher separately executes and calculates each step of the multi-modal data in the architecture to enhance the objectivity and accuracy of the experimental results. During the experiment, it was found that Mahā, which has limited computing power based on GPTs, frequently made errors in calculations during data analysis. Mahā only performs data in the experiment and records TP (true positives), TN (true negatives), FP (false positives), FN (false negatives) according to the confusion matrix principle. And the researcher used prompts to continuously revise Mahā's code errors during the calculation process, which also led to multiple experiments.

## V. RESULTS

Due to computer language conversion, each modality is input in the same way in Mahā, and the same evaluation indicators can be used.It provides objective evidence for evaluating the indicators of precision, recall, F1-score, specificity and accuracy. Taking into account ensuring accuracy, the researcher used the formula function of Microsoft Excel to calculate the data results of each indicator and display them in Table 1-3.

Specifically, Table 1 shows the results of running the Mahā simulation architecture to perform scene reasoning on visual synthetic data. The researcher recorded the seven modules represented by "Step 1" to "Step 7" respectively, and calculated the indicators based on the recorded TP, TN, FP, and FN.

Table 1: Visual performance metrics

| Mahā | Precision | Recall | F1-score | Specificity | Accuracy |
|---|---|---|---|---|---|
| Step 1 | 0.932 | 0.918 | 0.925 | 0.900 | 0.911 |
| Step 2 | 0.925 | 0.919 | 0.922 | 0.888 | 0.907 |
| Step 3 | 0.931 | 0.921 | 0.926 | 0.898 | 0.912 |
| Step 4 | 0.930 | 0.931 | 0.931 | 0.897 | 0.917 |
| Step 5 | 0.921 | 0.922 | 0.922 | 0.882 | 0.906 |
| Step 6 | 0.923 | 0.929 | 0.926 | 0.885 | 0.911 |
| Step 7 | 0.924 | 0.928 | 0.926 | 0.888 | 0.912 |

Table 2 presents the results of Mahā's computational metrics after execution on auditory synthesis data converted to computational language.

Table 2: Auditory performance metrics

| Mahā | Precision | Recall | F1-score | Specificity | Accuracy |
|---|---|---|---|---|---|
| Step 1 | 0.914 | 0.892 | 0.903 | 0.874 | 0.885 |
| Step 2 | 0.911 | 0.894 | 0.902 | 0.870 | 0.884 |
| Step 3 | 0.909 | 0.900 | 0.905 | 0.876 | 0.890 |
| Step 4 | 0.902 | 0.901 | 0.901 | 0.869 | 0.887 |
| Step 5 | 0.909 | 0.900 | 0.904 | 0.868 | 0.887 |
| Step 6 | 0.907 | 0.906 | 0.906 | 0.859 | 0.887 |
| Step 7 | 0.907 | 0.903 | 0.905 | 0.867 | 0.888 |

The following data analysis results in Table 3 reflect Mahā's implementation of tactile synthetic data converted into computational language.

Table 3: Tactile performance metrics

| Mahā | Precision | Recall | F1-score | Specificity | Accuracy |
|---|---|---|---|---|---|
| Step 1 | 0.892 | 0.879 | 0.885 | 0.858 | 0.870 |
| Step 2 | 0.889 | 0.891 | 0.890 | 0.851 | 0.874 |
| Step 3 | 0.893 | 0.888 | 0.890 | 0.863 | 0.877 |
| Step 4 | 0.895 | 0.886 | 0.891 | 0.863 | 0.876 |
| Step 5 | 0.890 | 0.876 | 0.883 | 0.856 | 0.868 |
| Step 6 | 0.891 | 0.886 | 0.888 | 0.855 | 0.873 |
| Step 7 | 0.892 | 0.879 | 0.886 | 0.856 | 0.869 |

## VI. DISCUSSION

As evidenced by findings, Mahā fully demonstrated rationality and feasibility in testing multi-scenario reasoning, although it failed to compare with data from companies such as Tesla, Nvidia, or Boston Dynamics. Table 1-3 shows that in the three modes of vision, hearing and touch, the five indicators of precision, recall, F1-score, specificity and accuracy of each module in the architecture are maintained at qualified and stable levels. Among them, the F1-score and accuracy of the attention-based prioritization and memory-augmented reasoning modules are particularly outstanding, which reflects their core role in multi-scenario reasoning. It proves the effectiveness of this architecture in solving the shortcomings of multi-modal understanding technology and thereby improving the autonomous cognitive capabilities of humanoid robots.

## VII. LIMITATIONS

This research focuses on the thinking, planning and action selection aspects of humanoid robots, which are abstract and logical in nature. This makes it difficult for the data synthesized by Gemini 2.0 Experimental Advanced to simulate physical dynamic interactions. The execution simulator Mahā may still be based on GPT4-Turbo in terms of computing power. Difficulty-differentiated prompts may affect the experimental results during execution. Synthetic data is created through generative code loops, which may be of variable quality when converted into computer language. In addition, the synthetic data used for simulation experiments does not take into account interfering factors in the physical scene such as reflection, noise, vibration, etc. The above situation means that the performance of the multi-scene reasoning architecture in real physical scene practice may be The results are slightly worse than those of the simulation experiment. Mahā has the potential to grow as OpenAI updates its ability to customize GPTs.

## VIII. CONCLUSION

Inspired by the principles of situated cognition theory, this research proposes a multi-scenario reasoning architecture in the field of humanoid robot thinking, planning and decision-making as an innovative solution to improve cognitive autonomy. By building and simulating experiments based on an architecture based on sparse attention and memory-augmented network, it proposes a new concept of "multi-scene reasoning" for the field to achieve multi-modal understanding of visual, auditory and tactile data. Experimental results show that the precision, recall, F1-score, specificity and accuracy of the above three modes displayed by the experimental simulator Mahā all remain at a stable level above 0.85. In particular, the attention-based prioritization and memory-augmented reasoning modules perform outstandingly. Despite the limitations of synthetic data and simulated environments, this finding confirms the effectiveness of multi-scenario reasoning in promoting cognitive autonomy in humanoid robots. It not only makes up for the technical shortcomings of multi-modal understanding in this field, but also provides feasible solutions and new development directions for future humanoid robots in thinking, planning and decision-making.

## ACKNOWLEDGMENT

I owe my deepest gratitude to Dr. Eva, my Doctor of Business Administration supervisor at UCSI University. Her encouragement and support have been essential in empowering me to explore artificial intelligence and autonomous robotics on my own, proving that passion surpasses background.I also want to thank Chuixin Wang, a senior AI engineer, for our insightful discussions over the past two years, which greatly accelerated my self-learning progress.Special thanks to Coursera and edX for providing free, high-quality educational resources, allowing me to build a solid knowledge foundation through their courses and certifications.

## References

[1] Burghart, C., Mikut, R., Stiefelhagen, R., Asfour, T., Holzapfel, H., Steinhaus, P., & Dillmann, R. (2005). A cognitive architecture for a humanoid robot: A first approach. In 5th IEEE-RAS International Conference on Humanoid Robots, 2005. (pp. 357-362). IEEE.

[2] Chen, K., Liu, J., Tong, Q., Cheng, X., & Zhang, J. (2024). Accurate Perception and Association of Objects for Humanoid Robots Under Dynamic Visual SLAM. International Journal of Humanoid Robotics, 21(3).

[3] Clavera, I., Rothfuss, J., Schulman, J., Fujita, Y., Asfour, T., & Abbeel, P. (2018). Model-based reinforcement learning via meta-policy optimization. In Conference on Robot Learning (pp. 617-629). PMLR.

[4] Condevaux, C., & Harispe, S. (2023). Lsg attention: Extrapolation of pretrained transformers to long sequences. In Pacific-Asia Conference on Knowledge Discovery and Data Mining (pp. 443-454). Cham: Springer Nature Switzerland.

[5] Dao, J., Duan, H., & Fern, A. (2024). Sim-to-real learning for humanoid box loco-manipulation. In 2024 IEEE International Conference on Robotics and Automation (ICRA) (pp. 16930-16936). IEEE.

[6] Datteri, E., & Schiaffonati, V. (2023). Computer simulations and surrogative reasoning for the design of new robots. Synthese, 202(1), 5.

[7] Duan, S., Shi, Q., & Wu, J. (2022). Multimodal Sensors and ML-Based Data Fusion for Advanced Robots. Advanced Intelligent Systems, 4(12), 2200213.

[8] Ekren, B. Y., Heragu, S. S., Krishnamurthy, A., & Malmborg, C. J. (2010). Simulation based experimental design to identify factors affecting performance of AVS/RS. Computers & Industrial Engineering, 58(1), 175-185.

[9] Fisher, M., Cardoso, R. C., Collins, E. C., Dadswell, C., Dennis, L. A., Dixon, C., ... & Webster, M. (2021). An overview of verification and validation challenges for inspection robots. Robotics, 10(2), 67.

[10] Guo, D., Li, Z., Khan, A. H., Feng, Q., & Cai, J. (2020). Repetitive motion planning of robotic manipulators with guaranteed precision. IEEE Transactions on Industrial Informatics, 17(1), 356-366.

[11] Jenlink, P. M., & Austin, F. S. (2013). Situated cognition theory. The handbook of educational theories, 185-198.

[12] Jouen, A. L., Ellmore, T. M., Madden, C. J., Pallier, C., Dominey, P. F., & Ventre-Dominey, J. (2015). Beyond the word and image: characteristics of a common meaning system for language and vision revealed by functional and structural imaging. NeuroImage, 106, 72-85.

[13] Katiyar, S., & Katiyar, K. (2021). Recent trends towards cognitive science: from robots to humanoids. In Cognitive Computing for Human-Robot Interaction (pp. 19-49). Academic Press.

[14] Kim, D., Choi, M., & Um, J. (2024). Digital twin for autonomous collaborative robot by using synthetic data and reinforcement learning. Robotics and Computer-Integrated Manufacturing, 85, 102632.

[15] Lou, C., Jia, Z., Zheng, Z., & Tu, K. (2024). Sparser is faster and less is more: Efficient sparse attention for long-range transformers. *arXiv preprint arXiv:2406.16747*.

[16] Luo, Y., Xu, H., Li, Y., Tian, Y., Darrell, T., & Ma, T. (2018). Algorithmic framework for model-based deep reinforcement learning with theoretical guarantees. arXiv preprint arXiv:1807.03858.

[17] Makoviychuk, V., Wawrzyniak, L., Guo, Y., Lu, M., Storey, K., Macklin, M., ... & State, G. (2021). Isaac gym: High performance gpu-based physics simulation for robot learning. arXiv preprint arXiv:2108.10470.

[18] Moerland, T. M., Broekens, J., Plaat, A., & Jonker, C. M. (2023). Model-based reinforcement learning: A survey. Foundations and Trends® in Machine Learning, 16(1), 1-118.

[19] Morici, J. F., Weisstaub, N. V., & Zold, C. L. (2022). Hippocampal-medial prefrontal cortex network dynamics predict performance during retrieval in a context-guided object memory task. Proceedings of the National Academy of Sciences, 119(20), e2203024119.

[20] Muthirayan, D., Nivison, S., & Khargonekar, P. P. (2020). Improved Attention Models for Memory Augmented Neural Network Adaptive Controllers. In 2020 American Control Conference (ACC) (pp. 639-646). IEEE.

[21] Navarro-Guerrero, N., Toprak, S., Josifovski, J., & Jamone, L. (2023). Visuo-haptic object perception for robots: an overview. Autonomous Robots, 47(4), 377-403.

[22] Nicolini, D., & Mengis, J. (2024). Toward a practice-theoretical view of the situated nature of attention. Strategic Organization, 22(1), 211-234.

[23] Ogunsina, M., Efunniyi, C. P., Osundare, O. S., Folorunsho, S. O., & Akwawa, L. A. (2024). Cognitive architectures for autonomous robots: Towards human-level autonomy and beyond.

[24] Pateria, S., Subagdja, B., Tan, A. H., & Quek, C. (2021). Hierarchical reinforcement learning: A comprehensive survey. ACM Computing Surveys (CSUR), 54(5), 1-35.

[25] Pawłowski, M., Wróblewska, A., & Sysko-Romańczuk, S. (2023). Effective techniques for multimodal data fusion: A comparative analysis. Sensors, 23(5), 2381.

[26] Pramanick, P., & Rossi, S. (2024). Multimodal Coherent Explanation Generation of Robot Failures. arXiv preprint arXiv:2410.00659.

[27] Saglam, A., & Papelis, Y. (2024). A Simulation-Based Approach for Evaluating Shared Control Algorithms for Mobile Robots. In 2024 Annual Modeling and Simulation Conference (ANNSIM) (pp. 1-13). IEEE.

[28] Song, L., Chen, Y., Yang, S., Ding, X., Ge, Y., Chen, Y. C., & Shan, Y. (2024). Low-Rank Approximation for Sparse Attention in Multi-Modal LLMs. In Proceedings of the IEEE/CVF Conference on Computer Vision and Pattern Recognition (pp. 13763-13773).

[29] Taniguchi, T., Murata, S., Suzuki, M., Ognibene, D., Lanillos, P., Ugur, E., ... & Pezzulo, G. (2023). World models and predictive coding for cognitive and developmental robotics: Frontiers and challenges. Advanced Robotics, 37(13), 780-806.

[30] Thagard, P. (2010). How brains make mental models. In Model-based reasoning in science and technology: Abduction, logic, and computational discovery (pp. 447-461). Berlin, Heidelberg: Springer Berlin Heidelberg.

[31] Tong, Y., Liu, H., & Zhang, Z. (2024). Advancements in humanoid robots: A comprehensive review and future prospects. IEEE/CAA Journal of Automatica Sinica, 11(2), 301-328.

[32] Wang, Y., Patel, S., Patel, D., & Wang, Y. (2003). A layered reference model of the brain. In The Second IEEE International Conference on Cognitive Informatics, 2003. Proceedings. (pp. 7-17). IEEE.

[33] Wu, X., Wang, G., & Shen, N. (2023). Research on obstacle avoidance optimization and path planning of autonomous vehicles based on attention mechanism combined with multimodal information decision-making thoughts of robots. Frontiers in Neurorobotics, 17, 1269447.

[34] Xiao, X., Liu, J., Wang, Z., Zhou, Y., Qi, Y., Cheng, Q., ... & Jiang, S. (2023). Robot learning in the era of foundation models: A survey. arXiv preprint arXiv:2311.14379.

[35] Ye, J., Jiao, W., Wang, X., Tu, Z., & Xiong, H. (2023). Cross-modality data augmentation for end-to-end sign language translation. arXiv preprint arXiv:2305.11096.

[36] Yuan, Y., Che, H., Qin, Y., Huang, B., Yin, Z. H., Lee, K. W., ... & Wang, X. (2024). Robot synesthesia: In-hand manipulation with visuotactile sensing. In 2024 IEEE International Conference on Robotics and Automation (ICRA) (pp. 6558-6565). IEEE.

[37] Zhang, J., Dean, E., & Ramirez-Amaro, K. (2024). Hierarchical Reinforcement Learning Based on Planning Operators. In 2024 IEEE 20th International Conference on Automation Science and Engineering (CASE) (pp. 2006-2012). IEEE.

[38] Zhao, Y., Zeng, Y., Long, Q., Wu, Y. N., & Zhu, S. C. (2023). Sim2Plan: Robot Motion Planning via Message Passing Between Simulation and Reality. In Proceedings of the Future Technologies Conference (pp. 29-42). Cham: Springer Nature Switzerland.